\PassOptionsToPackage{dvipsnames,table,xcdraw}{xcolor}
\documentclass[10pt,twocolumn,letterpaper]{article}

\usepackage[pagenumbers]{iccv} 

\definecolor{iccvblue}{rgb}{0.21,0.49,0.74}
\usepackage[pagebackref,breaklinks,colorlinks,allcolors=iccvblue]{hyperref}
\usepackage{stfloats}  
\usepackage{afterpage}  
\usepackage{graphicx}
\usepackage{caption}
\usepackage{subcaption} 
\usepackage{rotating} 
\usepackage{multirow} 
\usepackage{amssymb}
\usepackage[table,xcdraw]{xcolor}
\usepackage{wrapfig}
\usepackage{tabularx}
\usepackage{appendix} 

\def\paperID{5994} 
\def\confName{ICCV}
\def\confYear{2025}

\title{GAGS: Granularity-Aware Feature Distillation for Language Gaussian Splatting}

\author{Yuning Peng\textsuperscript{1,}\thanks{The first two authors contribute equally.},\ \ \ \
Haiping Wang\textsuperscript{1,$\ast$},\ \ \ \
Yuan Liu\textsuperscript{2,4},\ \ \
Chenglu Wen\textsuperscript{3},\ \ \ \
Zhen Dong\textsuperscript{1}\thanks{Corresponding Authors.},\ \ \ \
Bisheng Yang\textsuperscript{1,$\dagger$}\ \ \ \
\\
\textsuperscript{1}Wuhan University\ \ 
\textsuperscript{2}Hong Kong University of Science and Technology \ \
\textsuperscript{3}Xiamen University\\
\textsuperscript{4}Nanyang Technological University\ \
\\
{\tt\small yuningpeng@whu.edu.cn, hpwang@whu.edu.cn, yuanly@connect.hku.hk, clwen@xmu.edu.cn}\\
{\tt\small dongzhenwhu@whu.edu.cn, bshyang@whu.edu.cn}
}

\begin{document}

\twocolumn[{%
\renewcommand\twocolumn[1][]{#1}%
\maketitle

\centering

\includegraphics[width=\textwidth]{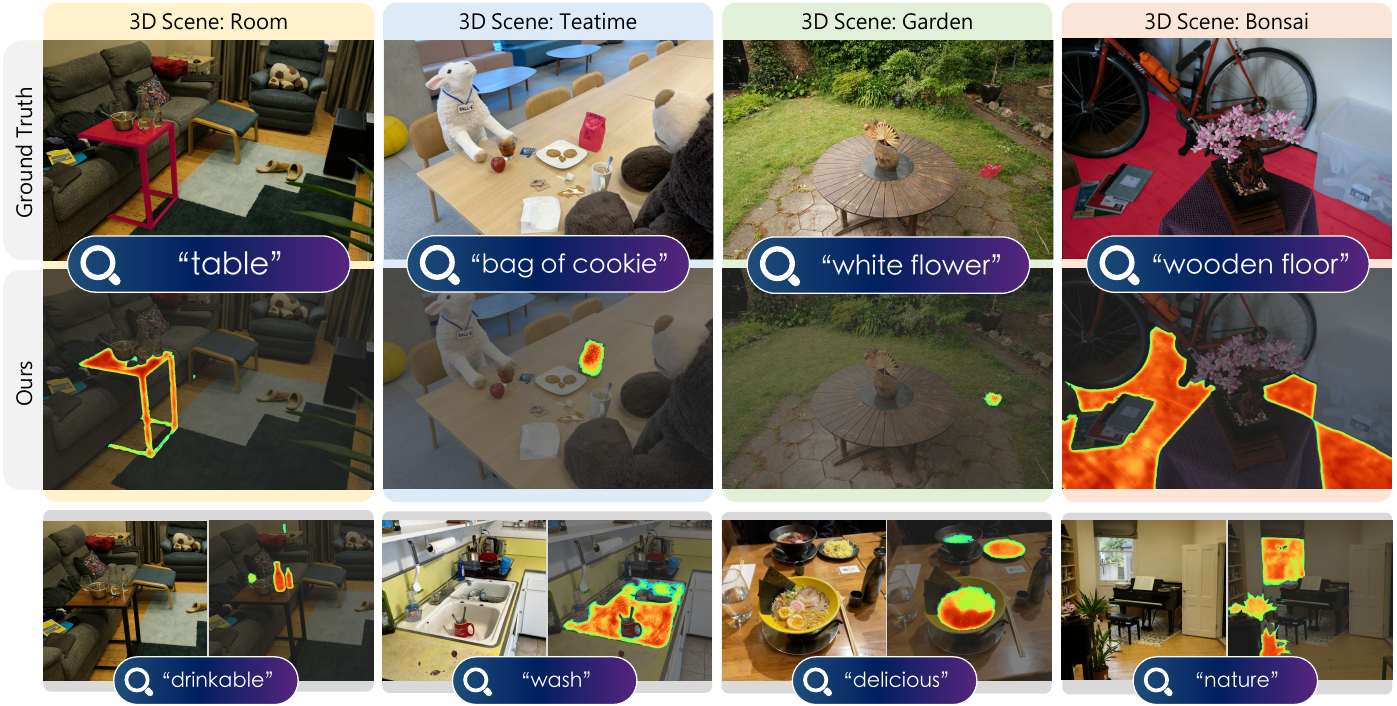}

\captionof{figure}{Given multiview images of a scene, GAGS learns a 3D Gaussian field associated with semantic features, which enables accurate open-vocabulary 3D visual grounding in the scene.} 
\label{fig:teaser}
\vspace{10pt}
}]

\begin{abstract}

\vspace{-10pt}
3D open-vocabulary scene understanding, which accurately perceives complex semantic properties of objects in space, has gained significant attention in recent years. 
In this paper, we propose GAGS, a framework that distills 2D CLIP features into 3D Gaussian splatting, enabling open-vocabulary queries for renderings on arbitrary viewpoints. 
The main challenge of distilling 2D features for 3D fields lies in the multiview inconsistency of extracted 2D features, which provides unstable supervision for the 3D feature field.
GAGS addresses this challenge with two novel strategies. First, GAGS associates the prompt point density of SAM with the camera distances, which significantly improves the multiview consistency of segmentation results. Second, GAGS further decodes a granularity factor to guide the distillation process and this granularity factor can be learned in a unsupervised manner to only select the multiview consistent 2D features in the distillation process. 
Experimental results on two datasets demonstrate significant performance and stability improvements of GAGS in visual grounding and semantic segmentation, with an inference speed 2$\times$ faster than baseline methods.
The code and additional results are available at \url{https://pz0826.github.io/GAGS-Webpage/}.
\vspace{-10pt}

\end{abstract}
     
\section{Introduction}
\label{sec:intro}


3D scene understanding is a fundamental task in computer vision and a critical challenge in fields like robotics~\citep{huang2023language_for_robot,shen2023language_manipulation,wang2024querybasedGS} and autonomous driving ~\citep{jatavallabhula2023conceptfusion,zheng2024genad}. Recent advances in artificial intelligence and deep learning have driven research on open-vocabulary scene understanding, enabling users to query scene models using natural language. This enhances the efficiency and intuitiveness of human-computer interaction, fostering closer integration between intelligent systems and human cognitive processes.

\begin{figure}[t]
    \centering
    \vspace{-10pt}
    \includegraphics[width=1.0\linewidth]{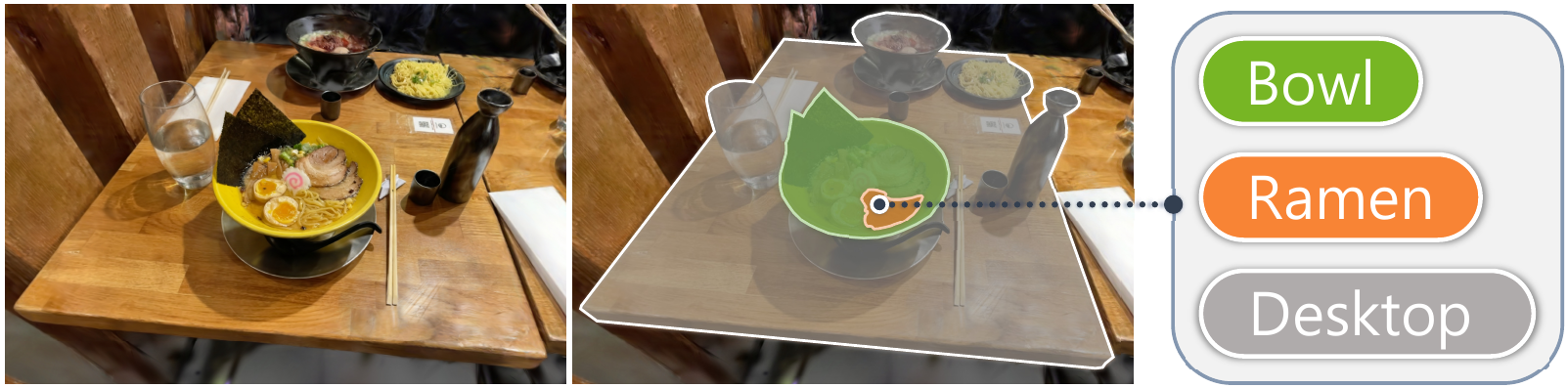}
    \vspace{-10pt}
    \caption{
    A 3D point shows different semantics at different scales.}
    \label{fig:intro_1}
    \vspace{-10pt}
\end{figure}

Due to the lack of large-scale, diverse 3D scene datasets with language annotations, current efforts focus on extending the knowledge of 2D vision-language models to 3D scenes. Early methods, such as OpenScene~\citep{peng2023openscene}, compute pixel-level embeddings using pre-trained segmentation models and project them onto 3D point clouds. Recent approaches like LERF~\citep{kerr2023lerf} employ NeRF~\citep{mildenhall2021nerf} to represent 3D scenes, directly integrating CLIP~\citep{radford2021CLIP} features into the scene modeling. Another recent work LangSplat~\citep{qin2024langsplat} extends the open-vocabulary feature learning to the 3D Gaussian Splatting (3DGS)~\citep{kerbl20233dgs} technique, which enables fast rendering and impressive performances of visual grounding with arbitrary text descriptions.

A noticeable challenge of learning such open-vocabulary features for 3D fields lies in the multiview inconsistency of the extracted 2D features. As shown in ~\cref{fig:intro_1}, given a specific 3D location, it can be described as ``desktop", ``bowl", and ``Ramen" at the same time according to the different granularities. 
Such semantic uncertainty often results in inconsistent feature extraction across varying viewpoints.
Brute-force training a 3D feature field from inconsistent 2D features leads to averaged and degenerated 3D features. 
LangSplat~\citep{qin2024langsplat} resolves this problem by learning three different feature fields corresponding to different segmentation granularity of SAM~\citep{kirillov2023sam}. However, learning three different fields increases both training memory and time cost and requires the query texts to be compared with three different features. 
FastLGS~\citep{ji2024fastlgs} accelerates open-vocabulary queries by indexing multi-view features of the same object into low-dimensional feature grids. However, it requires additional storage for high-dimensional CLIP features, which scales with the number of views.
N2F2~\citep{bhalgat2024n2f2} proposes to merge the features of different granularity by selecting the most activated features within a set of manually predefined descriptions. In this case, the performance is strongly related to the quality of these predefined descriptions. If the granularity of query descriptions differs from these predefined descriptions, the performance degenerates severely.

\begin{figure}[t]
    \centering
    \vspace{-10pt}
    \includegraphics[width=1.0\linewidth]{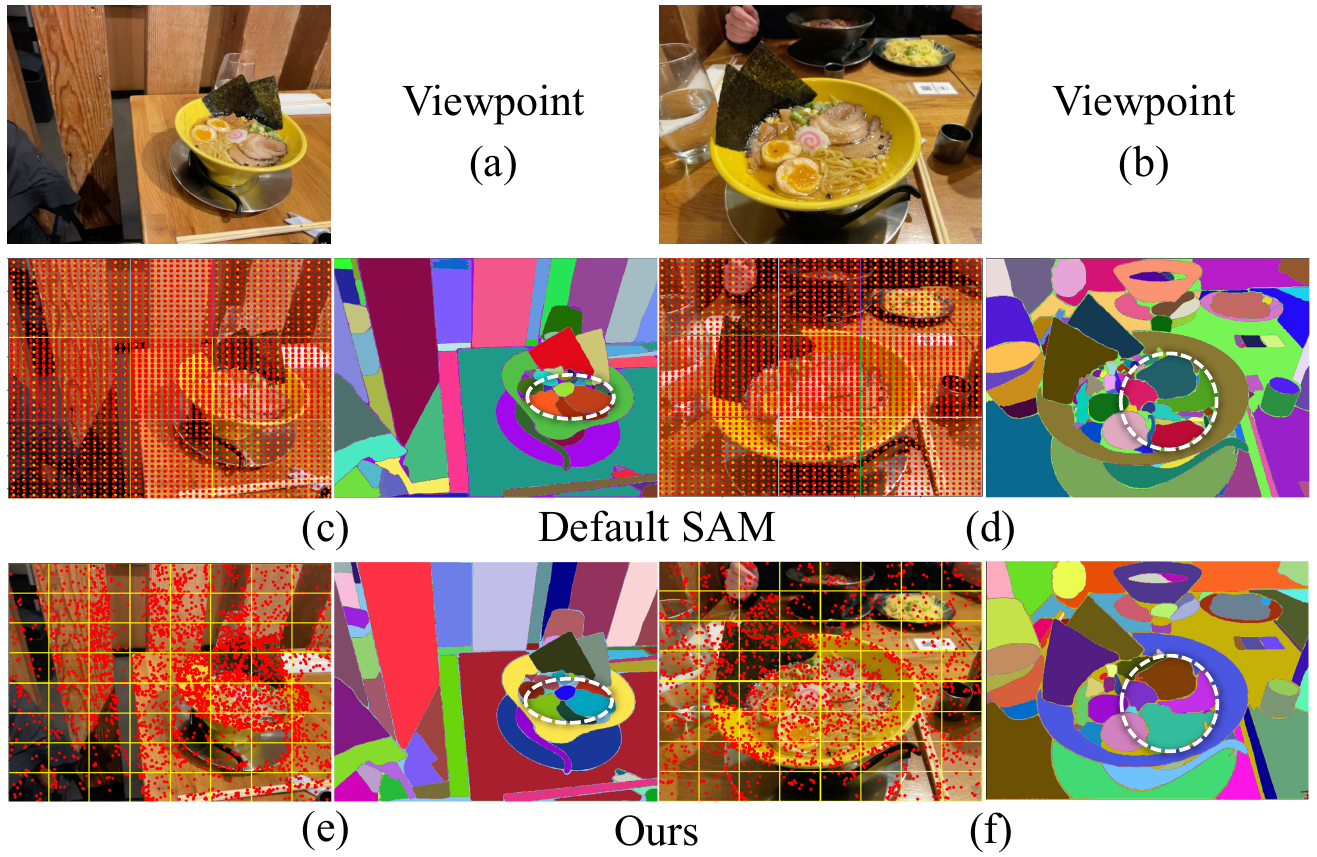}
    \vspace{-10pt}
    \caption{
    In comparison to vanilla SAM, we associate the prompt point density with distance to the object, which greatly enhances the consistency of multi-view segmentation.
    }
    \label{fig:intro_2}
    \vspace{-10pt}
\end{figure}

In this work, we introduce Granularity-Aware 3D Feature Learning for Gaussian Splatting (GAGS), a framework built on 3DGS that incorporates granularity-aware CLIP features in the reconstructed 3D Gaussian field. GAGS essentially consists of two strategies to overcome the inconsistency of multiview 2D features.

First, we improve the segmentation granularity to improve the consistency. 
In the CLIP feature extraction, we apply the SAM to segment images into regions for dense feature extraction, but its multi-solution nature causes viewpoint inconsistency. 
As shown in ~\cref{fig:intro_2}, we improve the consistency of SAM segmentation results among multiview images by introducing a granularity-aware prompting strategy. 
SAM relies on the prompt points to determine the segmentation granularity. 
Thus, we associate the prompt point density with the camera distance to the subject by utilizing the reconstructed 3D GS. Distant views receive dense prompts while nearby views use sparse prompts.
We found that this adaptive prompting effectively results in consistent segmentation across multiview images and further improves the CLIP feature consistency.

\begin{figure*}
    \centering
    \vspace{-20pt}
    \includegraphics[width=0.9\textwidth]{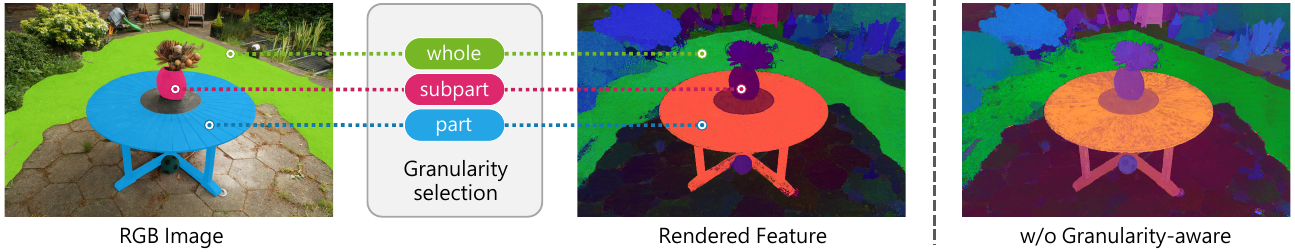}
    \vspace{-5pt}
    \caption{\textit{Our granularity-select feature learning.}
    Our method leverages the inherent consistency of 3D Gaussian splatting to perform granularity-aware feature distillation, enhancing the stability and accuracy of learned object features.}
    \label{fig:intro_3}
    \vspace{-10pt}
\end{figure*}

Second, we incorporate a granularity decoder on the learned 3D feature to decode a granularity factor as shown in ~\cref{fig:intro_3}, which helps the distillation process only learn the multiview consistent 2D features while neglecting inconsistent ones. On every image, we simultaneously extract three levels of 2D CLIP features, i.e. sub-parts, parts, and objects. Then, the decoded granularity factor will be incorporated in the distillation loss to select the CLIP feature of a specific granularity for training. Note that the granularity scale is automatically learned in the distillation process without additional supervision.

We conduct experiments on the LERF~\cite{kerr2023lerf} dataset and a self-annotated Mip-NeRF-360 dataset for evaluation. Experimental results demonstrate that our method outperforms baseline methods in the open-vocabulary localization and semantic segmentation tasks with a 2$\times$ improvement in query speed.

\section{Related work}
\label{sec:relatedwork}

\subsection{Point-based Radiance Field}

Radiance fields have long been used to represent 3D scenes \citep{gortler1996lumigraph,levoy1996light}. Recently, neural radiance field methods, exemplified by NeRF \citep{mildenhall2021nerf}, have achieved breakthroughs in rendering quality, bringing renewed attention to this field. Trained NeRF models leverage neural networks to learn complete scene representation, enabling high-quality novel view rendering. However, due to the reliance on ray sampling and deep MLP architectures, NeRF suffers from slow training and rendering speeds. While various approaches \citep{muller2022instant, fridovich2022plenoxels} have attempted to improve these aspects, they still struggle to balance speed and quality. Additionally, NeRF's implicit representation makes it difficult to directly obtain the scene geometry. Methods such as NeuS \citep{wang2021neus} and VolSDF \citep{yariv2021volumeSDF} address this by introducing Signed Distance Functions (SDF) to represent geometric surfaces, deploying new volumetric rendering techniques to learn the SDF representation. 3D Gaussian Splatting \citep{kerbl20233dgs} introduces a novel solution as a point-based radiance field method, representing scenes using discrete 3D Gaussian distributions. It replaces expensive random-sampling-based volumetric rendering with $\alpha$-blending, significantly boosting rendering speed. Additionally, the explicitly stored Gaussian kernels allow direct retrieval of scene geometry after training. Leveraging these advantages, we adopt 3DGS as the framework and extend it to support open-vocabulary 3D scene understanding.

\subsection{SAM}

As one of the most impressive foundational vision models, Meta’s SAM \citep{kirillov2023sam} has demonstrated exceptional zero-shot 2D segmentation capabilities. SAM supports flexible prompts, allowing it to generate multi-level masks for target objects based on inputs such as points, bounding boxes, and masks. Additionally, SAM offers the capability to automatically segment the entire image into multi-granularity masks, including whole, subpart and part. Numerous methods have since emerged to extend SAM’s capabilities to 3D space. Anything-3D \citep{shen2023anything3d} elevates SAM's segmentation to 3D, while Part123 \citep{liu2024part123} uses SAM’s segmentation masks to reconstruct 3D models with high-quality segmented parts. 
Feature 3DGS \citep{zhou2024feature3dgs} and Gaussian Grouping \citep{ye2023gsgrouping} integrate SAM's features and object segmentation results at pre-selected single-granularity into 3D Gaussians, achieving high-quality segmentation of novel views and 3D scenes. Semantic Gaussian \citep{guo2024semanticgaussian} utilizes various prompts to obtain instance-level segmentations for feature extraction. 


Considering that SAM can produce multi-granularity masks focusing on objects at different scales, some methods \citep{kim2024garfield,ying2024omniseg3d,liang2024supergseg} integrates these masks into scene representations jointly or separately. 
However, the above methods, whether single-granularity or multi-granularity, do not address the issue of multi-view mask conflicts. In contrast, we enhance the consistency and reliability of multi-view masks in SAM segmentation and mask integration process.


\subsection{Open-vocabulary Scene Understanding}

Scene understanding is a basic task in the field of computer vision. With the rapid advancement of deep learning, numerous methods \citep{wu20153dshapenet,chang2017matterport3d,chen2020scanrefer,wang2021synthesizing} have made significant progress across various subtasks of scene understanding. However, the limited availability of 3D training data has long posed a challenge for achieving comprehensive 3D scene understanding. 
Some methods \citep{wu2025opengaussian,ye2023gsgrouping} attempt to align the outcomes of 2D grounding models~\citep{ren2024groundedsam} with 3D category-agnostic grouping approaches, thereby indirectly achieving open-vocabulary understanding. 
Vision foundation models like CLIP \citep{radford2021CLIP} have opened new avenues for open-vocabulary scene understanding. Due to the image-aligned nature of CLIP features, subsequent works such as CLIP2Scene \citep{chen2023clip2scene} and Openscene \citep{peng2023openscene} directly leverage CLIP-based 2D scene understanding models \citep{dong2023maskclip,li2022lseg} to obtain dense features with CLIP semantics. Approaches like \citep{zhang2023clipfo3d,liu20233dovs,kerr2023lerf,zuo2024fmgs,shi2024legaussian} generate dense semantic features by performing multi-level image cropping and feature fusion, while often incorporating pixel-aligned feature supervision, such as DINO \citep{caron2021dino}, to address the blurriness in semantic boundary. Some other researches \citep{kobayashi2022F3RM,hong20233dllm,huang2023vlmap,qin2024langsplat} utilize pre-trained image segmentation model~\citep{li2022lseg,cheng2022mask2former} to obtain semantically meaningful object-level patches, enabling more accurate scene understanding.

\begin{figure*}[t]
    \centering
    \vspace{-20pt}
    \includegraphics[width=\linewidth]{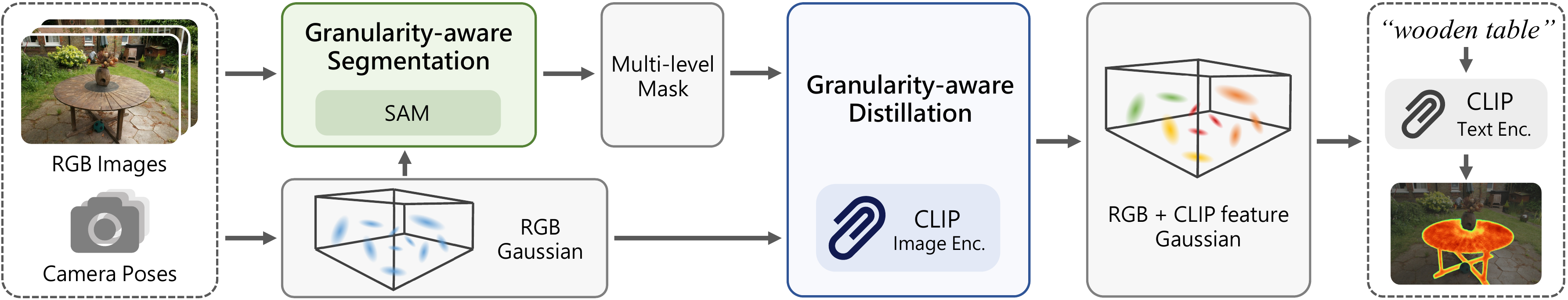}
    \vspace{-12pt}
    \caption{\textit{Pipeline of our method. }
    Given a set of images with camera poses, our method first uses 3D Gaussian Splatting to reconstruct the scene's geometric representation, then utilizes it for granularity-aware segmentation and CLIP feature distillation. The finally output 3D feature field supports open-vocabulary queries.
    }
    \vspace{-10pt}
    \label{fig:pipline}
\end{figure*}

\section{Method}
\label{sec:method}
Given multiple input images with known camera poses, our target is to learn a 3D feature field represented by 3D Gaussians with associated feature vectors. Then, we can render arbitrary novel-view images from this 3D feature field and also use text prompts to query the 3D feature field for a range of downstream tasks, such as object localization or semantic segmentation. 

An overview of our method is provided in \cref{fig:pipline}. Inspired by LangSplat~\citep{qin2024langsplat}, We first train a 3D Gaussian field from the input posed multiview images. Then, for every input image, we apply the SAM to get a set of segmented regions. Each region is cropped and fed into the CLIP image encoder to get a CLIP feature, which is regarded as the feature for all pixels in the segmented region. Finally, we associate a trainable feature vector on each Gaussian and apply the splatting technique to render a feature map from the 3D Gaussian field. By minimizing the difference between the rendered feature map and the extracted CLIP feature map, we train these associated feature vectors on Gaussians. The main challenge lies in extracting and learning multiview-consistent CLIP features. In GAGS, we propose the following Granularity-aware Segmentation and Granularity-aware Distillation to address this challenge.

\subsection{Granularity-aware Segmentation}
The extracted CLIP features are strongly related to the region size of the segmentation results produced by SAM, different segmentation granularity leads to totally different CLIP features. However, since the SAM is applied to every input image separately and the same object may show different sizes on different viewpoints, the segmentation results of SAM are already multiview inconsistent, resulting in inconsistent CLIP features. For example, the ramen bowl in \cref{fig:intro_2}(c)(d) corresponds to different segmentation granularities across various viewpoints. We propose a granularity-aware segmentation to improve the multiview consistency of segmentation results.

\textbf{Observations}. The segmentation granularity of SAM is controlled by the prompt point density. In the ramen bowl region of \cref{fig:intro_2}(d)(f), dense prompt points usually lead to small segmentation regions while sparse prompt points result in large regions. 
Another observation is that objects far from the camera appear smaller in the image, requiring smaller segmentation masks, while nearby objects appear larger and require larger masks. In this case, prompt point density becomes depth-dependent: higher density is needed for distant viewpoints to segment smaller objects, while lower density suffices for nearby viewpoints to segment larger objects.

\begin{figure}[t]
    \centering
    \includegraphics[width=\linewidth]{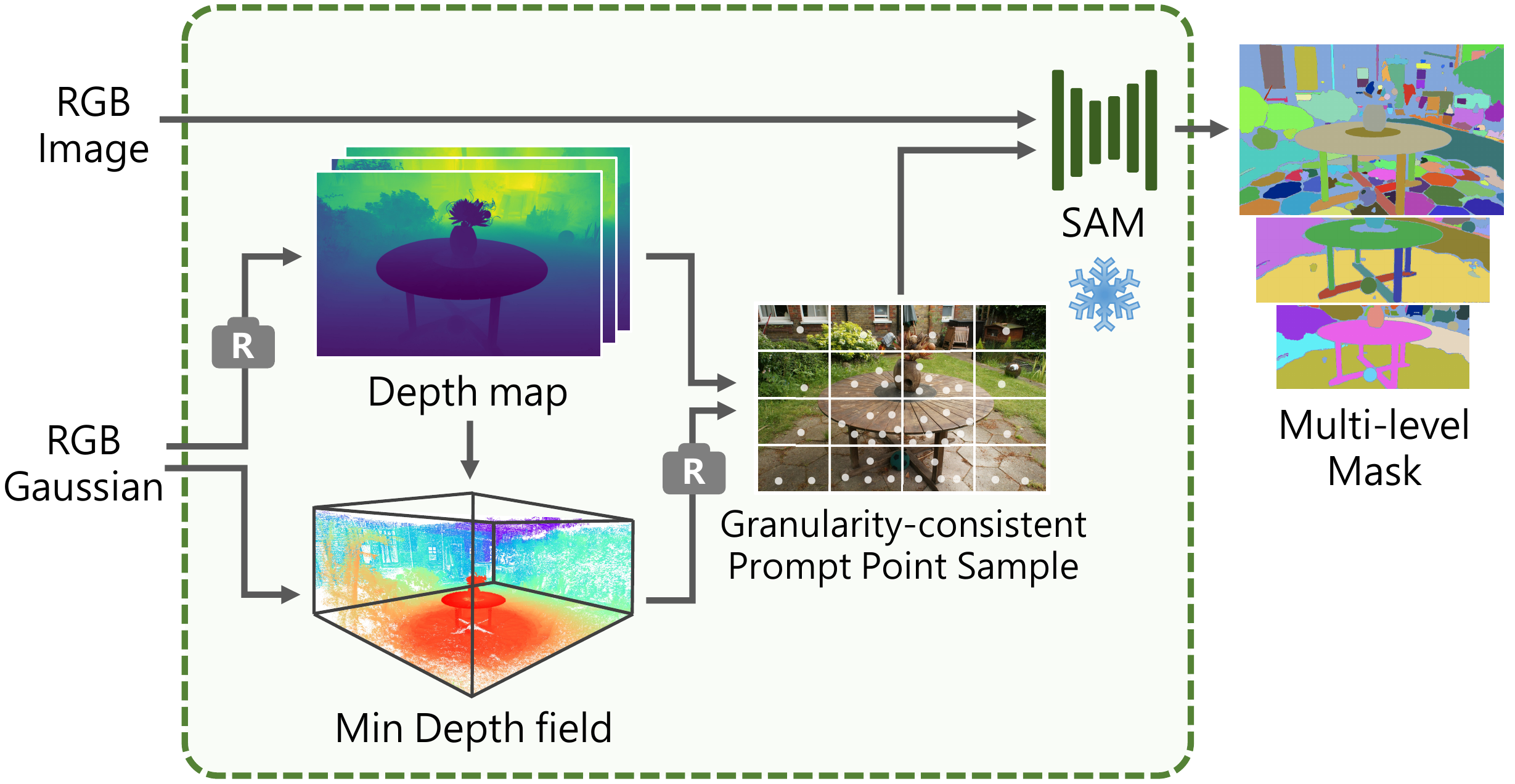}
    \vspace{-12pt}
    \caption{
    Our granularity-aware segmentation module utilizes the Gaussian kernel point cloud and camera poses to guide SAM and CLIP in producing multi-view consistent semantic features.
    }
    \vspace{-10pt}
    \label{fig:pipline_segmentation}
\end{figure}



\textbf{Granularity-aware prompt point density}. We compute the prompt point density from the depth maps as follows. As illustrated in \cref{fig:pipline_segmentation}, Given an input image, we begin by dividing it into a grid of patches and our target is to determine how many sample points should lie in each patch. Since the rendered depth determines the density, we first render a depth map $D$ from the 3D Gaussian field. 
Then, we back-project each depth map pixel to the 3D Gaussians and find the minimum depth value of each Gaussian after occlusion determination, denoted by $MD$, which maps a pixel to the minimum visible depth across all views. Then, the number of prompt points $n_{P}$ for a specific patch $P$ is computed by
\begin{equation} 
n_{P}=\frac{1}{|P|}\sum_{p\in P}\frac{D^2(p)}{MD^2(p)}\cdot n,
\label{eq:density}
\end{equation}
where $|P|$ is the pixel number in the patch, where $|P|$ is the pixel number in the patch, $p\in P$ is a pixel, $MD^2(p)$ is the squared minimum depth value of pixel $p$ while $D^2(p)$ is the squared depth value of pixel $p$, and $n$ is a predefined prompt point number.

 
\textbf{Explanation of \cref{eq:density}}. 
The ratio $\frac{D^2(p)}{MD^2(p)}$ controls the prompt point density on a specific view. When this view is the one with the smallest viewing distance, the ratio is 1.0, which means we sample $n$ points on the nearest view. While, for a far-away viewpoint, we use a larger prompt point density because the object would be smaller on these views, which can be seen in~\cref{fig:intro_2}(e)(f). In this way, ~\cref{eq:density} enables granularity-invariant prompting for the SAM model when segmenting the same object from different viewpoints, which increases the multi-view consistency of the segmentation results. 

Moreover, after determining the number of prompt points per patch, the density distribution of visible gaussians is utilized to guide local prompt point sampling. 
Please refer to the supplementary material for further details.

\begin{figure*}[t]
    \centering
    \vspace{-20pt}
    \includegraphics[width=0.9\linewidth]{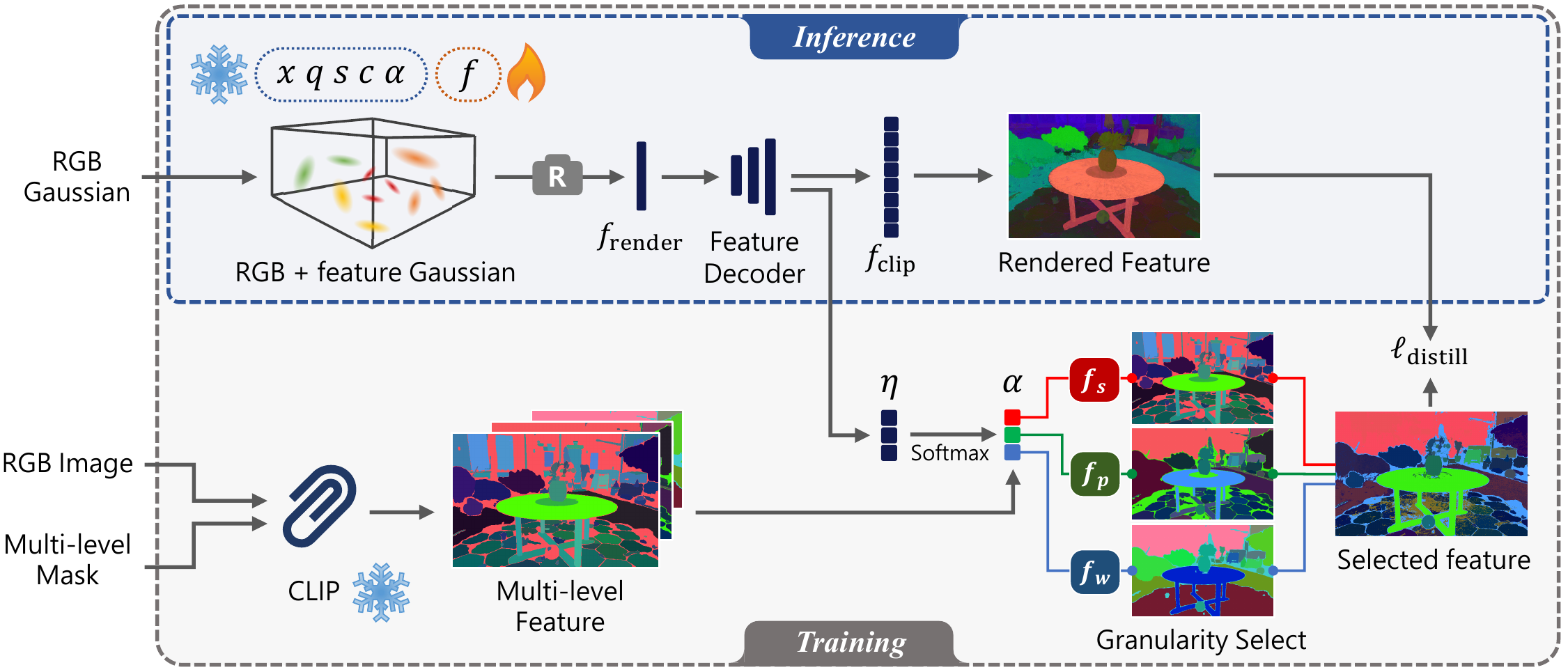}
    \vspace{-5pt}
    \caption{
    Our granularity-aware language field distillation module introduces optimizable low-dimensional semantic features into the pre-trained Gaussian scene. The rendered semantic features are fed into the granularity decoding head, which selects an appropriate scale for each pixel. These selected features are then used to compute loss with the output full-dim semantics of the feature decoder for optimization.
    }
    \vspace{-12pt}
    \label{fig:pipline_distillation}
\end{figure*}

\subsection{Granularity-aware Distillation}

While the above granularity-aware segmentation improves the consistency, SAM's prediction still exists with randomness and leads to inconsistency. Thus, we further propose a novel granularity-aware distillation to autonomously select the granularity with the best multi-view consistency during optimization. Given the granularity-aware prompt points, SAM returns three levels of segmentation results $m_{s}$, $m_{p}$ and $m_{w}$ corresponding to ``sub-part", ``part", and ``whole". We then extract CLIP feature maps $f_{s}$, $f_{p}$ and $f_{w}$ on all three levels and design a strategy to let the distillation automatically select the most multiview-consistent features.

\textbf{Decoding granularity factor}. 
Given the 3D Gaussian field associated with trainable feature vectors, we apply the splatting technique to render a feature map $f_{\text{render}}$ for a training viewpoint. We adopt a feature decoder to decode a granularity factor and a predicted CLIP feature from the rendered feature vectors by
\begin{equation}
    f_{\text{clip}},\eta=\mathcal{D}(f_{\text{render}}),
\end{equation}
where $f_{\text{clip}}$ is the predicted CLIP feature, $\eta\in \mathbb{R}^3$ is the predicted granularity factor, and $\mathcal{D}$ means the MLP-based decoder shared among all viewpoints. 


\textbf{Weighted distillation loss}. We convert the granularity factor into weight by applying the softmax operator between three scores $\alpha=\text{softmax}(\eta)$. Then, we will train with a weighted L2 loss for distillation
\begin{equation}
    \ell_{\text{distill}} = \sum_{n\in \{s,p,w\}} \alpha_n \|f_{\text{clip}}-f_{n} \|_2^2,
\end{equation}
where $n$ means the name, $s,p,w$ correspond to ``sub-part", ``part", and ``whole" respectively, and $\alpha_n\in [0,1]$ means the weight for the current level. We do not directly supervise $\alpha_n$ but let the optimization process automatically select the best weight to learn the granularity of the feature. 

\textbf{Entropy regularization}. To further encourage the weight to converge to a single scale instead of learning an averaged feature, we adopt an entropy regularization term
\begin{equation}
    \ell_{\text{entropy}} = -\sum_{n\in \{s,p,w\}} \alpha_n\log \alpha_n,
\end{equation}
which encourages the weight to be either 0 or 1. By adopting the granularity-aware distillation, GAGS only learns one 3D feature field instead of 3 feature fields in LangSplat~\cite{qin2024langsplat}, resulting in a more compact representation. The predicted granularity factor enables adaptively adjusting the granularity for different views and selecting the most multiview-consistent one.

\textbf{Region-aware weighted distillation}. In our experiments, we observed that due to perspective variations and intrinsic object scale differences, different objects occupy varying proportions within the same image. During loss calculation, objects with a larger region in the image contribute more significantly to the loss, thereby dominating the optimization direction of semantic features. This imbalance may hinder feature learning for smaller objects. To address this issue, we propose a region-aware distillation loss, normalizing the loss by the region size of each object to ensure that all objects contribute equally to the total loss
\begin{equation}
    \ell_{\text{r-distill}}=\beta_{r}{\ell_{\text{distill}}},\beta_{r}=\frac{\sum_{i=1}^{n_{r}}{S}({R}_i)}{n_{r}\cdot{{S}({R})}},
\end{equation}
where $\beta_{r}$ denotes the region-aware factor, with $n_{r}$ as the number of objects in the image and ${S}({R})$ as the area of each region. 
However, region-aware strategy is non-trivial.
The core challenge here is constructing a mask that can simultaneously segment both large and small objects for balanced supervision, where SAM masks at any single granularity  fails to achieve this as the whole scale mask $m_w$ focuses on large objects, while the part $m_p$ and subpart $m_s$ masks emphasize small objects. Therefore, we propose integrating SAM masks at multiple granularities to adaptively select the appropriate scale for objects of different sizes.
Specifically, we construct $m_f$, where each pixel selects the SAM granularity with the highest weight, i.e., $m_f = m_{\underset{i \in \{s,p,w\}}{\arg\max}(\alpha_i)}$. We then cluster pixels in $m_f$ to objects for computing $S(R)$. Further details are provided in the supplementary material.




\textbf{Feature consistency loss}. 
Inspired by the process of contrastive learning, we further introduce a feature consistency loss to encourage the features inside the same segmentation region to be consistent with each other
\begin{equation}
    \ell_\text{cons}=\sum _{i=1}^{n_r}\sum _{p \in R_i}^{}\frac{(f_{clip}^p-\overline{f_{R_i}})^2}{S(R_i)}
\end{equation}
where $f_{clip}^p$ and $\overline{f_{R_i}}$ represent the $f_{\text{clip}}$ of pixel $p$ and the average $f_{\text{clip}}$ of the $i$-th region. By pulling together features of the same object region across each viewpoint, we ultimately enhance the feature consistency for the 3D Gaussians belonging to the same object.

In summary, the optimization loss $\mathcal{L}$ is
\begin{equation}
\mathcal{L}=\ell_\text{r-distill}+\lambda_\text{entropy}\ell_\text{entropy}+\lambda_\text{cons}\ell_\text{cons}
\end{equation}


\begin{table*}[ht]
    \centering
    \vspace{-20pt}
    \begin{tabular}{l|ccccc|ccccc}
        \multirow{2}{*}{\textbf{Method}} & \multicolumn{5}{c|}{\textbf{LERF}} & \multicolumn{5}{c}{\textbf{Mip-NeRF360}} \\ 
        ~ & \textit{Ramen} & \textit{Teatime} & \textit{Kitchen} & \textit{Figurines} & \textit{Overall} & \textit{Room} & \textit{Counter} & \textit{Garden} & \textit{Bonsai} & \textit{Overall} \\ \hline\hline
        GS-Grouping & 32.39 & 69.49 & 50.00 & 44.64 & \cellcolor[HTML]{EFEFEF}49.13 & \underline{79.31} & 56.76 & 57.14 & 66.67 & \cellcolor[HTML]{EFEFEF}64.97 \\ 
        LEGaussian & \textbf{69.01} & \underline{79.66} & 63.64 & 57.14 & \cellcolor[HTML]{EFEFEF}67.36 & 55.17 & 72.97 & \underline{71.42} & 61.11 & \cellcolor[HTML]{EFEFEF}65.17 \\
        GOI & 56.33 & 67.80 & 68.18 & 44.64 & \cellcolor[HTML]{EFEFEF}59.24 & 68.97 & 67.57 & 66.67 & \underline{72.22} & \cellcolor[HTML]{EFEFEF}68.86 \\
        Langsplat & \underline{63.38} & \textbf{88.14} & \underline{81.82} & \underline{76.79} & \cellcolor[HTML]{EFEFEF}\underline{77.53} & 75.86 & \underline{91.89} & 52.38 & \underline{72.22} & \cellcolor[HTML]{EFEFEF}\underline{73.09} \\  \hline
        GAGS(\textbf{Ours}) & \textbf{69.01} & \textbf{88.14} & \textbf{90.91} & \textbf{78.57} & \cellcolor[HTML]{EFEFEF}\textbf{81.66} & \textbf{93.10} & \textbf{97.30} & \textbf{80.95} & \textbf{83.33} & \cellcolor[HTML]{EFEFEF}\textbf{88.67} \\ 
    \end{tabular}
    \vspace{-5pt}
    \caption{\textit{Quantitative comparisons of 3D object location on the LERF and Mip-NeRF 360 dataset.} We report the mean accuracy(\%↑).}
    \vspace{-5pt}
    \label{mACC_comparsion}
\end{table*}

\begin{table*}[ht]
    \centering
    \begin{tabular}{l|ccccc|ccccc}
        \multirow{2}{*}{\textbf{Method}} & \multicolumn{5}{c|}{\textbf{LERF}} & \multicolumn{5}{c}{\textbf{Mip-NeRF360}} \\ 
        ~ & \textit{Ramen} & \textit{Teatime} & \textit{Kitchen} & \textit{Figurines} & \textit{Overall} & \textit{Room} & \textit{Counter} & \textit{Garden} & \textit{Bonsai} & \textit{Overall} \\ \hline\hline
        LEGaussian & 20.17 & 32.29 & 22.3 & 23.41 & \cellcolor[HTML]{EFEFEF}24.54 & 25.49 & 35.26 & 33.18 & 22.29 & \cellcolor[HTML]{EFEFEF}29.06 \\
        GS-Grouping & 26.39 & 53.97 & 31.33 & 34.55 & \cellcolor[HTML]{EFEFEF}36.56 & 54.38 & 47.67 & 40.36 & 54.13 & \cellcolor[HTML]{EFEFEF}49.14 \\ 
        GOI & 33.69 & 55.79 & \underline{54.46} & 23.85 & \cellcolor[HTML]{EFEFEF}41.95 & \underline{60.26} & 46.64 & \underline{59.75} & \underline{67.26} & \cellcolor[HTML]{EFEFEF}\underline{58.48} \\
        Langsplat & \underline{45.82} & \textbf{62.47} & 43.39 & \underline{51.10} & \cellcolor[HTML]{EFEFEF}\underline{50.70} & 51.22 & \textbf{64.69} & 49.83 & 52.96 & \cellcolor[HTML]{EFEFEF}54.68 \\  \hline
        GAGS(\textbf{Ours}) & \textbf{46.81} & \underline{60.29} & \textbf{55.8} & \textbf{53.59} & \cellcolor[HTML]{EFEFEF}\textbf{54.12} & \textbf{65.16} & \underline{61.05} & \textbf{61.22} & \textbf{70.52} & \cellcolor[HTML]{EFEFEF}\textbf{64.49} \\ 
    \end{tabular}
    \vspace{-5pt}
    \caption{\textit{Quantitative comparisons of 3D semantic segmentation on the LERF and Mip-NeRF 360 dataset.} We report the mean IoU(\%↑).}
    \vspace{-10pt}
    \label{mIOU_comparsion}
\end{table*}

\begin{figure*}[ht]
    \centering
    \vspace{-25pt}
    \includegraphics[width=\textwidth]{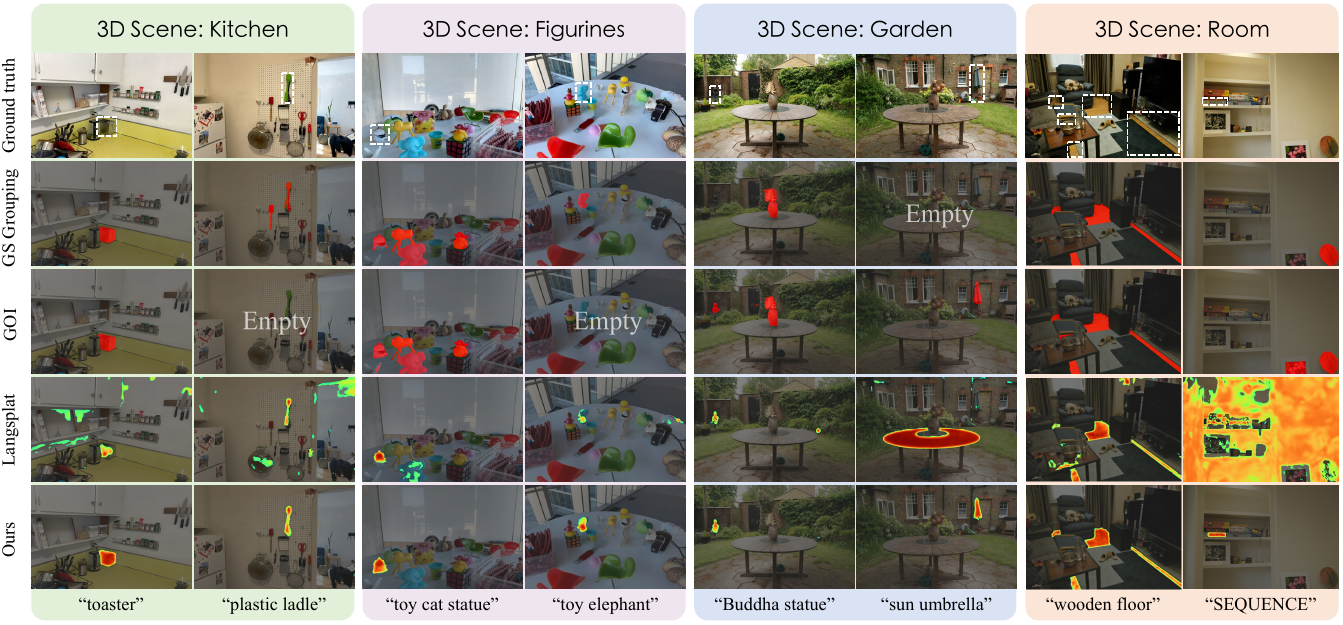}
    \caption{\textit{The relevance visualization results for open-vocabulary queries.} Each row from top to bottom represents Ground Truth, Gaussian Grouping, GOI, Langsplat, and our method. Below each column is the corresponding input text description.}
    \label{fig:main comparison}
    \vspace{-10pt}
\end{figure*}

\section{Experiments}
\subsection{Experimental Protocol}

\textbf{Datasets}. We use the LERF dataset~\citep{kerr2023lerf} and a self-annotated Mip-NeRF-360~\citep{barron2022mipnerf360} dataset for evaluation.  The LERF~\citep{kerr2023lerf} dataset comprises 3700+ phone-captured images from 14 different scenes. LangSplat~\cite{qin2024langsplat} provides text descriptions and corresponding multi-view segmentation masks within four scenes: ``ramen'', ``waldo\_kitchen'', ``teatime'', and ``figurines'', aiming to evaluate text-based object localization and segmentation. The Mip-NeRF-360~\citep{barron2022mipnerf360} dataset provides multi-view images for 9 indoor and outdoor scenes. Following LangSplat~\cite{qin2024langsplat}, we annotated four of the most complex scenes: ``room'' , ``counters'', ``garden'' and ``bonsai''. For each scene, we provided query texts for multiple types of objects along with the corresponding multi-view segmentation masks.

\textbf{Baselines.} We adopt recent relevant works including GS-Grouping~\cite{ye2023gsgrouping}, LEGaussian~\cite{shi2024legaussian}, GOI~\cite{qu2024goi}, and LangSplat~\cite{qin2024langsplat} as our baseline methods. 
We employ GSplat~\cite{ye2024gsplat} to construct initial Gaussian fields and evaluate all methods in the same setting for fair comparisons.


\textbf{Metrics.} We follow LERF~\citep{kerr2023lerf} and LangSplat~\citep{qin2024langsplat} to evaluate text-based 3D localization and segmentation accuracy on multi-view images. 
For localization, we evaluate the mean accuracy (mAcc) of the predicted locations falling within the ground truth bounding boxes. 
For segmentation, we assess the mean Intersection over Union (mIoU) between the predicted and the ground truth masks.

\textbf{Implementation Details.} We utilized the SAM ViT-H~\citep{dosovitskiy2020vit} and OpenCLIP ViT-B/16~\citep{cherti2023openclip} for segmentation and CLIP feature extraction. 
All experiments are conducted on a single RTX-4090 GPU. More details on implementation are provided in the supplementary material.

\subsection{Comparison with Baseline Methods}

In ~\cref{fig:main comparison}, we visualize the segmentation results of the baseline methods and our method. \cref{mACC_comparsion} and \cref{mIOU_comparsion} report the localization and segmentation accuracy of these methods on the LERF and Mip-NeRF-360 datasets. All performances reported were re-evaluated and the metrics of Langsplat has some differences from the original paper, we provide further analysis in the supplementary material.

As shown in ~\cref{fig:main comparison}, GS Grouping~\cite{ye2023gsgrouping} achieves sharp segmentation results by fusing multi-view SAM segments. However, segmentation masks for the same object across multiple views might lack consistency, failing to support and merge with each other. The incorrect fusion of segmentation masks can lead to detection failures, which is especially noticeable on small objects.
GOI~\cite{qu2024goi} uses detection results from the Grounding-SAM~\cite{liu2023groundingdino} to fine-tune a semantic-space hyperplane for selecting the final segment. However, Grounding-SAM also tends to fail with small and long-tail objects, causing errors in hyperplane optimization and leading to missed detections.
LangSplat~\cite{qin2024langsplat} embeds multi-level CLIP features within its Gaussian field, achieving better localization accuracy by avoiding hard fusion. However, the multi-view CLIP features often lack consistency, which introduces substantial noise into the feature field, yielding erroneous or noisy segmentation outcomes.

In contrast, our method encourages the distillation of multi-view consistent CLIP features into the Gaussian field, avoiding noise from conflicting multi-view features, which leads to a 4.1\% improvement in localization mAcc and 3.4\% in segmentation mIoU on the LERF dataset. 
Moreover, the proposed Granularity-aware Distillation can adaptively fuses multi-granularity features, thus ensuring effective multi-scale object localization. Consequently, on the Mip-NeRF-360 dataset, which includes objects of varying scales, we achieved a more significant 15.5\% mAcc and 6.0\% mIoU improvement over the baseline.

\begin{table}[t]
    \centering
    \resizebox{\linewidth}{!}{
    \begin{tabular}{l|cccc|ccccc}
        \multirow{2}{*}{\textbf{Method}} & \multicolumn{4}{c|}{\textbf{Training (min)}} & \multicolumn{5}{c}{\textbf{Inference (s)}} \\ 
        ~ & Prep. & GS & Lang. & Total & IE & Rend. & Pred. & OSH & Total \\ \hline\hline
        GS-Grouping & 20 & 10 & 40 & \cellcolor[HTML]{EFEFEF}70 & 66 & \raisebox{-0.5ex}{\textasciitilde}1 & \raisebox{-0.5ex}{\textasciitilde}1 & - & \cellcolor[HTML]{EFEFEF}68 \\ 
        GOI & 10 & 10 & 10 & \cellcolor[HTML]{EFEFEF}\textbf{30} & - & \raisebox{-0.5ex}{\textasciitilde}1 & \raisebox{-0.5ex}{\textasciitilde}1 & 2342 & \cellcolor[HTML]{EFEFEF}2344 \\
        Langsplat & 50 & 10 & 30 & \cellcolor[HTML]{EFEFEF}90 & - & 15 & 34 & - & \cellcolor[HTML]{EFEFEF}49 \\  \hline
        GAGS(\textbf{Ours}) & 25 & 10 & 50 & \cellcolor[HTML]{EFEFEF}85 & - & 13 & 11 & - & \cellcolor[HTML]{EFEFEF}\textbf{24} \\ 
    \end{tabular}
    }
    \vspace{-5pt}
    \caption{\textit{Time evaluation for each step in training and prediction on the scene “ramen” of the LERF dataset.} During training, “Prep.” refers to the preprocessing stage, while “GS” and “Lang.” represent the training stage of RGB and feature Gaussian fields, respectively. During inference, “IE” and “OSH” denotes obtaining the target's Identity Encoding and the optimization process of the Optimizable Semantic-space Hyperplane, while “Rend.” and “Pred.” refer to rendering features and predicting target mask.}
    \label{time}
    \vspace{-10pt}
\end{table}

\textbf{Runtime analysis}.
We report the runtime of baselines and GAGS in \cref{time}.
Although GOI~\cite{qu2024goi} achieved the shortest training time, it requires fine-tuning a high-dimensional hyperplane during testing to segment the target, resulting in an inference time nearly 100 times that of GAGS.
When comparing with LangSplat~\cite{qin2024langsplat}, GAGS achieves a similar training speed. However, LangSplat requires rendering multi-granularity feature maps to compare and segment the target, while GAGS adaptively fuses multi-granularity features, necessitating only a single feature map rendering and comparison. Thus, GAGS is two times faster than LangSplat in inference.

\begin{figure*}[ht]
    \centering
    \vspace{-20pt}
    \includegraphics[width=\linewidth]{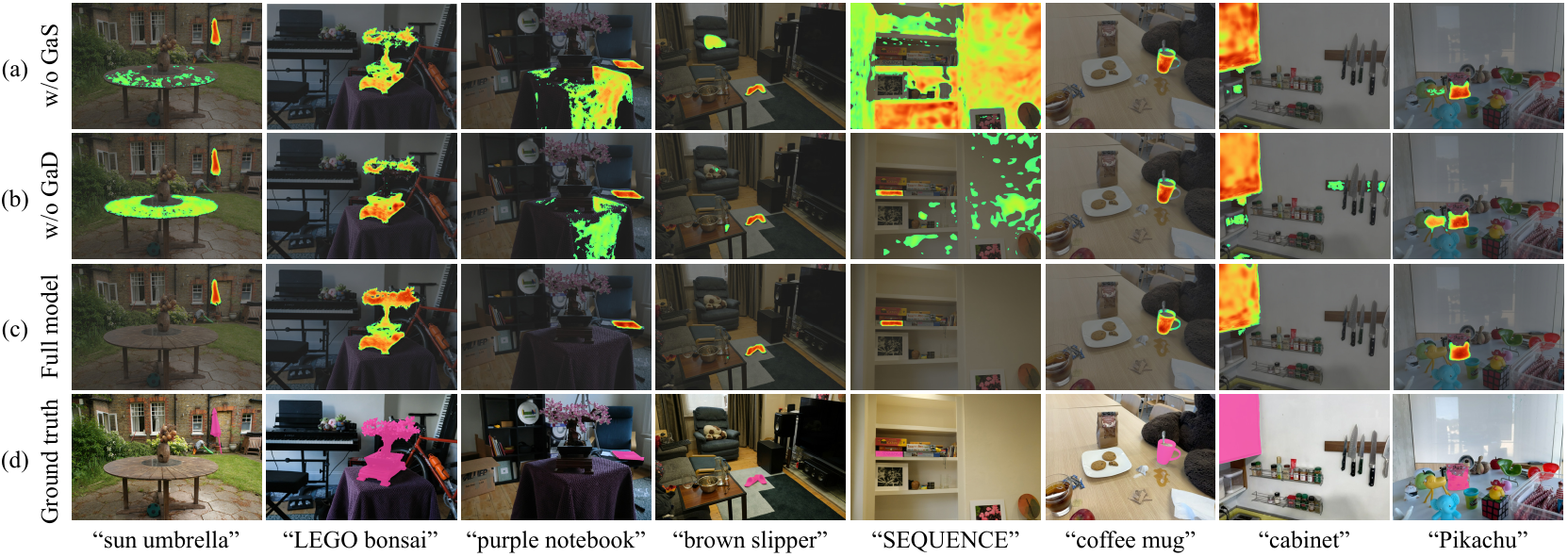}
    \vspace{-15pt}
    \caption{\textit{Qualitative results of the ablation study.} Below each column is the corresponding input text description.
    }
    \vspace{-12pt}
    \label{fig:compare_ablation}
\end{figure*}

\begin{figure}[t]
    \centering
    \includegraphics[width=\linewidth]{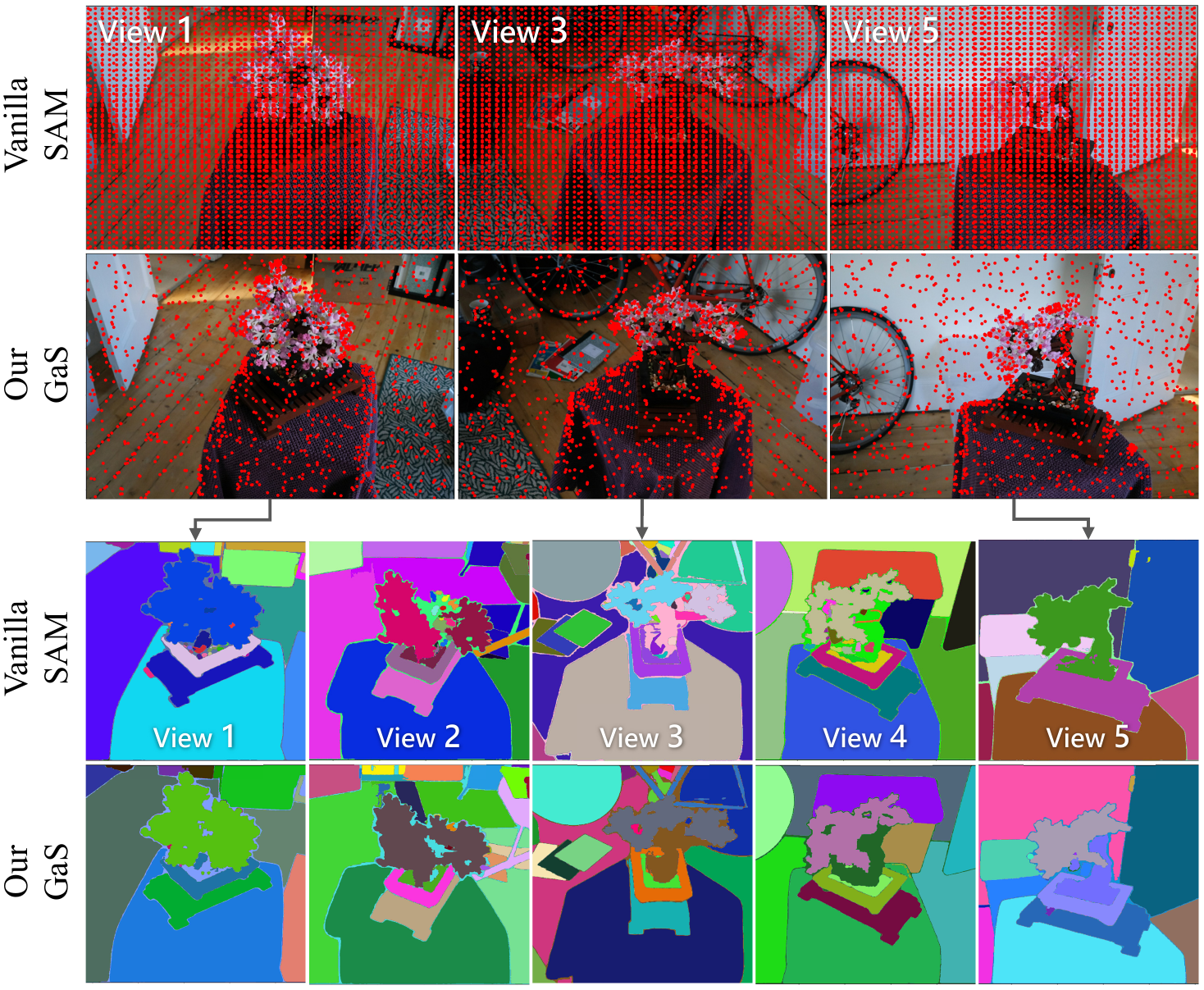}
    \vspace{-15pt}
    \caption{\textit{Visualization of prompt points and segmentation results.} Our method achieves superior segmentation consistency with only \raisebox{-0.5ex}\textasciitilde20\% of the prompt point count used in the vanilla SAM.}
    \label{fig:GaS_vis}
    \vspace{-10pt}
\end{figure}

\begin{figure}[t]
    \centering
    \includegraphics[width=\linewidth]{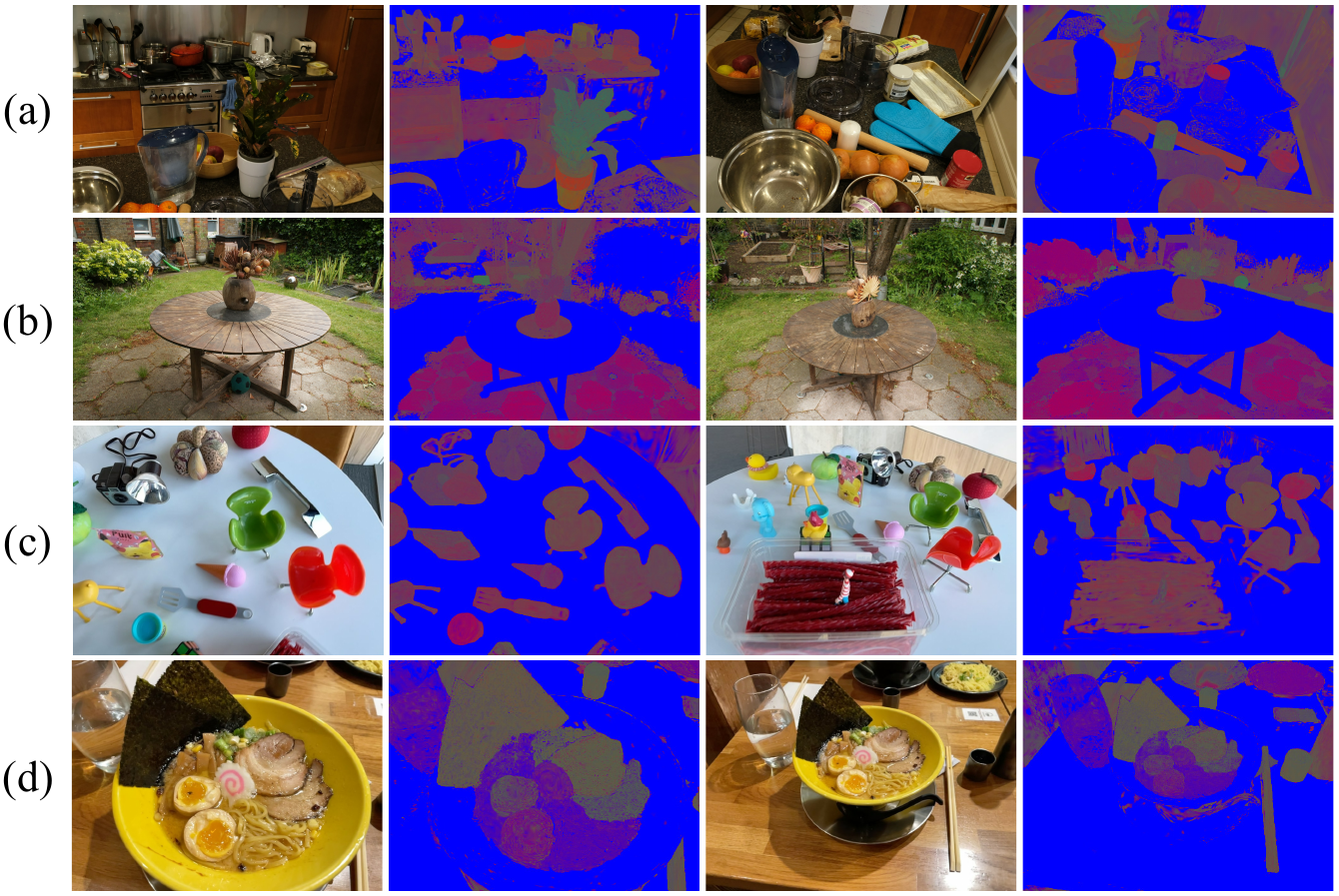}
    \vspace{-15pt}
    \caption{\textit{Visualization of learned granularity weights $\alpha$.} The RGB channels correspond to $\alpha_s$, $\alpha_p$, and $\alpha_w$, respectively.}
    \label{fig:Weight_vis}
    \vspace{-15pt}
\end{figure}

\subsection{Ablation Studies}
In~\cref{ablation_studies} and~\cref{fig:compare_ablation}, we conduct both qualitative and quantitative ablation evaluations of GAGS on the Mip-NeRF-360 dataset. 
Model 1 of~\cref{ablation_studies} is the baseline model, where we use uniform prompt points for SAM segmentation. 
We then average the CLIP features, calculated as $(f_s+f_p+f_w)/3$, and distill the normalized feature into Gaussian field.
Model 2 incorporate the proposed Granularity-aware Distillation (GaD), which boosts the localization mAcc and segmentation mIoU by 9.4\% and 7.2\%, respectively.
Further incorporating the proposed Granularity-aware Segmentation (GaS) in Model 3 brings an additional 2.4\% mAcc and 3.5\% mIoU improvement. 
As shown in Fig.~\ref{fig:GaS_vis}, GaS achieves better multiview segmentation consistency by prompting SAM with fewer but proper seeds.
The results in~\cref{fig:compare_ablation}(a) and (c) also shows that GaS can mitigate noise issues from over-segmentation of nearby objects and low-texture areas, as well as under-segmentation of distant objects.

\begin{table}[t]
    \centering
    \footnotesize
    \begin{tabular}{ccc|cc}
        \multicolumn{3}{c|}{\textbf{Setting}} & \multicolumn{2}{c}{\textbf{Performance}} \\ 
        ID & GaS & Distill. & \textbf{mIoU(\%)} & \textbf{mAcc(\%)}  \\
        \hline\hline
        1 & ~ & Avg. & 53.72 & 76.84 \\
        2 & ~ & GaD & 60.99 & 86.27 \\ 
        \rowcolor[HTML]{EFEFEF}
        3 & \textbf{\checkmark} & \textbf{GaD} & \textbf{64.49}  &\textbf{88.67} \\ 
        4 & \checkmark & $f_s$ & 47.78 & 77.32 \\
        5 & \checkmark & $f_p$ & 61.68 & 85.74 \\
        6 & \checkmark & $f_w$ & 58.63 & 78.51 \\
        7 & \checkmark & Avg. & 59.45 & 84.88 \\
    \end{tabular}
    \vspace{-5pt}
    \caption{\textit{Ablation Studies on the Mip-NeRF-360 dataset.} 
    ``Distill.'' denotes the feature distillation strategy. For ``Distill.'', in addition to distilling the averaged features across all granularities, we also conducted experiments for each individual granularity feature.}
    \vspace{-10pt}
    \label{ablation_studies}
\end{table}

Moreover, in Models 4-6, we distill CLIP features using single SAM segmentation granularity. Among them, the part granularity yields the best results, while the subpart granularity performed the worst because the over-segmentation makes CLIP feature extraction unreliable and leads to multi-view conflicts. Model 7, which simply averages the CLIP features from all three SAM granularities, shows no improvement. In contrast, Model 3 with GaD can adaptively select reliable and multi-view consistent features.
As shown in Fig.~\ref{fig:Weight_vis}, GaD facilitates the model in selecting features of corresponding granularities for objects of varying scales, ensuring the consistency and robustness of object feature learning.
Finally, model 3 achieves optimal performance. The comparison between panels (b) and (c) in~\cref{fig:compare_ablation} also demonstrates that GaD significantly improves segmentation accuracy by distilling multi-view consistent features at the appropriate granularity for each object.

\section{Conclusion}

In this paper, we propose GAGS, a granularity-aware 3D Gaussian Splatting framework for efficient and accurate open-vocabulary queries. By introducing a granularity-aware segmentation and feature distillation strategy during the learning of 3D feature field, GAGS is able to learn clear and multiview consistent semantic features. It avoids the high query cost of rendering multiple language fields and mitigates the impact of outliers. Experimental results across several datasets demonstrate that our method outperforms existing approaches in both performance and speed.

\appendix
{
    \small
    \bibliographystyle{ieeenat_fullname}
    \bibliography{main}
}

\clearpage

\begin{figure*}[t]
    \centering
    \vspace{-10pt}
    \includegraphics[width=\linewidth]{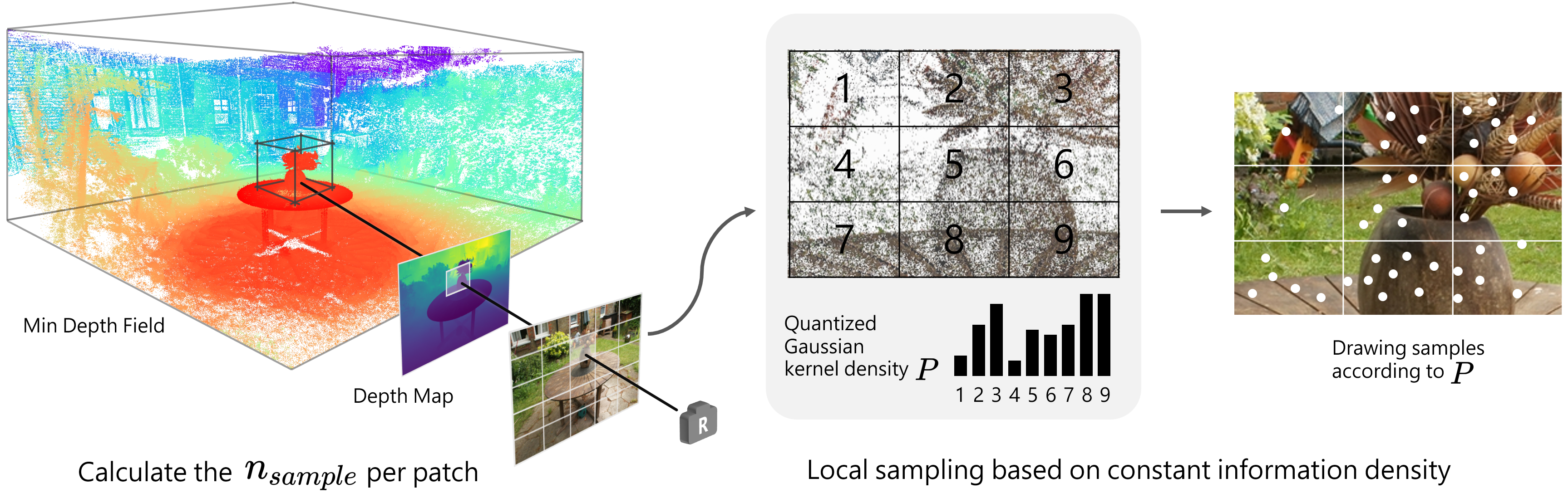}
    \vspace{-12pt}
    \caption{\textit{Granularity-aware segmentation.}
    For an input image, we divide it into a series of patches and calculate the number of prompt points for each patch using the depth-aware density. Next, we quantify the local Gaussian density into a discrete probability distribution to guide the sampling process within each patch.
    }
    \vspace{-10pt}
    \label{fig:sample_method}
\end{figure*}


\section{More Implementation Details}

\textbf{Prompt Point Sampling}. After obtaining the number of prompt points $n_P$ for a patch, we do not simply employ a uniform or random sampling approach, as information is not evenly distributed in each patch. This is evident from the distribution of Gaussian in the scene — regions with rich information contain more Gaussians, while information-sparse regions contain fewer. Therefore, we use the density of visible Gaussian kernels to represent information density, guiding the local sampling of prompt points accordingly. Considering that the distribution of Gaussian kernels is fixed in the scene, this sampling approach helps enhance the multi-view consistency of prompt points. Specifically, as shown on the right side of \cref{fig:sample_method}, each patch is further divided into a series of sub-patches, where we calculate the number of pixels with corresponding Gaussian kernels. These counts are then converted into discrete probability distributions, which are used to sample prompt points within each patch.
We found that it can balance segmentation granularity between information-rich and sparse regions, thus ensuring comprehensive segmentation of each object in the image.

\begin{figure}[t]
    \centering
    \includegraphics[width=\linewidth]{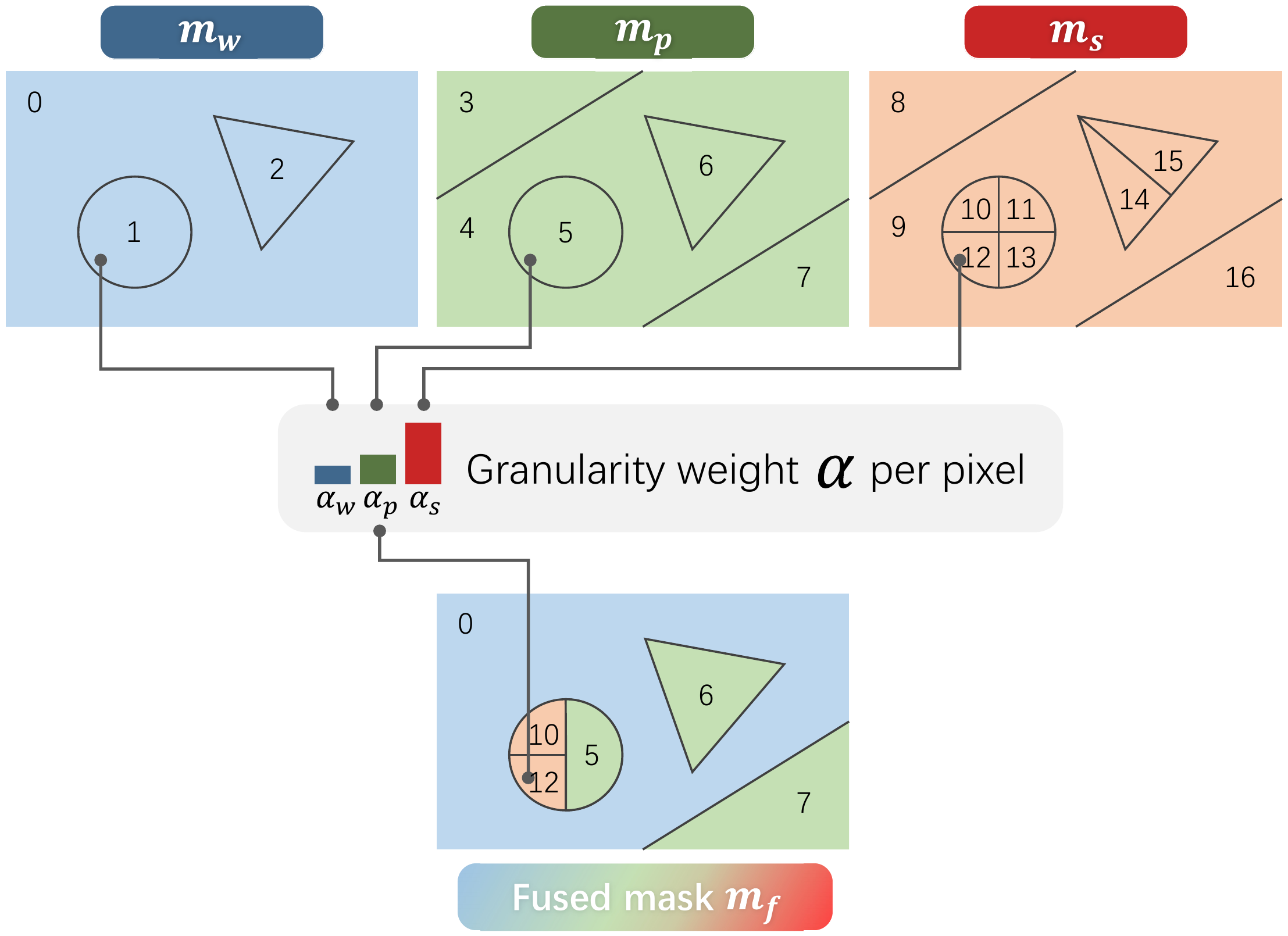}
    \vspace{-12pt}
    \caption{\textit{Mask fusion process in the training phase.}
    }
    \vspace{-10pt}
    \label{fig:mask_fusion}
\end{figure}

\textbf{Granularity-aware Mask Fusion for Training}. 
We perform mask fusion to automatically generate appropriate masks for objects at various granularities during each training forward pass. The fused mask $m_f$ is used for computing both the Region-aware Weighted Distillation and the Feature Consistency Loss. As illustrated in \cref{fig:mask_fusion}, for each view, we first assign a unique ID to each object region in all levels of SAM masks $\{m_i \mid i \in s,p,w\}$. Then, for each pixel in $m_f$, we determine its granularity level using the current $\alpha_i$ and assign it the corresponding ID from the mask at that level, which can be represented as 
\begin{equation}
    m_f = m_{\underset{i \in \{s,p,w\}}{\arg\max}(\alpha_i)}.
\end{equation}
Then, pixels sharing the same ID in $m_f$ are treated as a single object, contributing equally for the distillation loss, and being regularized internally by the Feature Consistency Loss during optimization.

\textbf{3D Object Localization \& Segmentation}. 
We follow LERF~\citep{kerr2023lerf} and LangSplat~\citep{qin2024langsplat} to evaluate text-based 3D localization and segmentation accuracy on multi-view images. 
GS-Grouping~\citep{ye2023gsgrouping} can directly obtain segmentation masks based on the mask ID identified by grounded SAM~\citep{ren2024groundedsam}, with the localization result determined by the center of the segmentation mask. 
For other baselines and our method, we first render multi-view feature maps. Then, LEGaussian~\citep{shi2024legaussian} selects pixels with high cosine similarity to text features as segmentation results. GOI~\citep{qu2024goi} further enhances the segmentation mask through an optimizable semantic-space hyperplane. We use feature maps smoothed by mean filtering and select the pixel with the highest relevancy to text features as the localization result.

LangSplat renders and smooths CLIP features $f_{clip}$ at three granularities, then separately computes pixel-level relevancy maps with the query text CLIP features $f_{text}$ by  
\begin{equation}
\min_{i}~\frac{exp(f_{clip} \cdot f_{text})}{\exp(f_{clip} \cdot f_{canon}^i)+\exp(f_{clip} \cdot f_{text})},
\end{equation}
where $f_{text}$ means the CLIP feature of query text, $f_{canon}^i, i=1,...,4$ represents CLIP features of four predefined keywords~\citep{kerr2023lerf}, specifically ``\textit{object}'', ``\textit{things}'', ``\textit{stuff}'' and ``\textit{texture}''. 
Afterward, LangSplat selects the pixel with the highest relevancy as the localization result, using a threshold of 0.4 to filter the normalized relevancy map as the segmentation result. Our method is similar to LangSplat but uses only one feature map for relevancy calculation and further segmentation.

We used the default parameters provided in the released code for all baseline methods during training and inference. For each method, we conducted three independent training and inference runs and selecting the best result among them. However, in our tests, LangSplat's performance metrics showed some differences compared to the original paper, primarily in terms of object localization performance. This discrepancy could be due to the following reasons:

\uppercase\expandafter{\romannumeral1}. Their results may have used hyperparameters or methods for computing predictions and metrics that differ from the released version. To this end, We emailed the authors to inquire about this, but unfortunately received no response.

\uppercase\expandafter{\romannumeral2}. Their results might be based on a larger number of repeated experiments, and reporting only the best-case results. Considering the inherent instability in Gaussian feature field learning, this assumption seems possible.

\begin{figure*}[t]
    \centering
    \vspace{-10pt}
    \includegraphics[width=\linewidth]{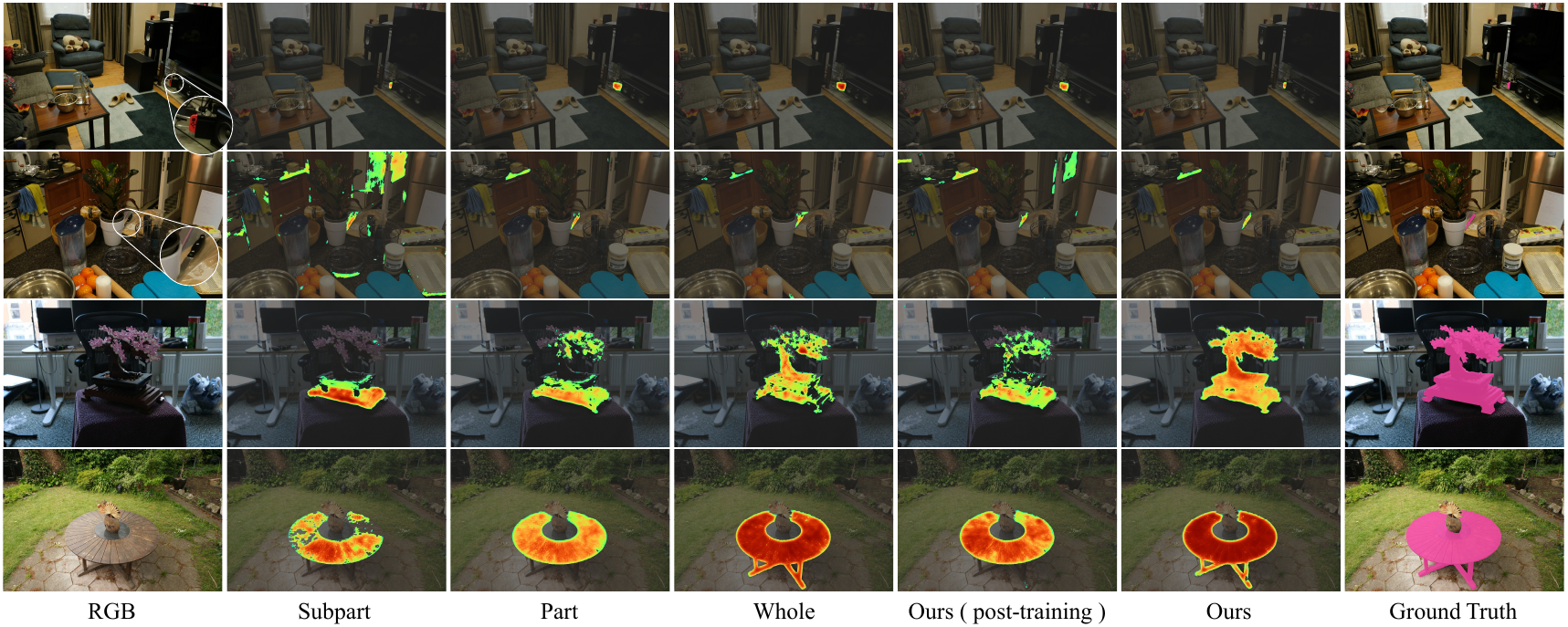}
    \vspace{-12pt}
    \caption{Comparison of the semantic segmentation result of N2F2-style post-training granularity-aware strategy and our granularity-aware strategy optimized during training, using the query ``red Nintendo Switch joy-con controller'' ``knife'' ``LEGO bonsai'' and ``wooden table''.
    }
    \vspace{-10pt}
    \label{fig:ablation_N2F2}
\end{figure*}

\section{More Ablation Study}
\textbf{Weighted fusion in N2F2~\cite{bhalgat2024n2f2} v.s. our granularity-aware strategy}. 
We show the comparison in \cref{tab:ablation_mIOU_N2F2}.
Since N2F2~\cite{bhalgat2024n2f2} does not release their codes, we reimplement it by performing a weighted fusion of multi-granularity features after training based on their relevance to a set of predefined descriptive phrases. Specifically, for the Gaussian $\Theta_i$ in the feature field, the weight $w_i \in \mathcal{R}^3$ can be obtained by 
\begin{equation}
w_i=\underset{n \in \{s,p,w\}}{\text{Softmax}}\left(\max_k\left(f_n\right)^\top f_k^{canon}\right),
\end{equation}
where $s,p,w$ correspond to ``sub-part", ``part", and ``whole" respectively, $\left \{f_k^{canon}\right \}$ is the set of the CLIP embeddings of the predefined phrases, specifically ``\textit{object}'', ``\textit{thing}'', ``\textit{stuff}'', ``\textit{part}'' and ``\textit{texture}''. We observed that the post-training approach to compute granularity fusion weights may result in inappropriate feature blending in some cases, leading to incorrect responses for queries with granularities that have weak relevance to predefined phrases. As shown in \cref{fig:ablation_N2F2}, the post-training approach does not produce the correct scale combination weights in these cases. In contrast, our method selects the feature with appropriate granularity based on multi-view consistency during the optimization process.

\begin{table}[t]
    \centering
    \begin{tabular}{l|cc}
        \multirow{2}{*}{\textbf{Scene}} & \multicolumn{2}{c}{\textbf{Setting}}  \\ 
        ~ & \textit{N2F2~\cite{bhalgat2024n2f2}} & \textit{Ours}  \\ \hline\hline
        \textit{Room} & 89.66 & 93.10 \\
        \textit{Counter} & 94.59 & 97.3  \\ 
        \textit{Garden} & 66.67 & 80.95 \\
        \textit{Bonsai} & 83.33 & 83.33 \\ 
        \rowcolor[HTML]{EFEFEF}
        \textit{Overall} & 83.56 & 88.67 \\ 
    \end{tabular}
    \vspace{-5pt}
    \caption{Quantitative comparison of the two granularity-aware approaches for 3D object localization on the Mip-NeRF-360 dataset. We report the mean accuracy(\%↑). ``N2F2~\cite{bhalgat2024n2f2}.'' refers to the approach that applies weighted fusion of multi-granularity features in the post-training stage.}
    \vspace{-10pt}
    \label{tab:ablation_mIOU_N2F2}
\end{table}

\begin{table}[t]
    \centering
    \begin{tabular}{ccc|cc} 
        \multicolumn{3}{c|}{\textbf{Setting}} & \multicolumn{2}{c}{\textbf{Performance}} \\ 
        RAD & EL & FCL & \textbf{mIoU(\%)} & \textbf{mAcc(\%)}  \\
        \hline\hline
        ~ & ~ & ~ & 55.14 & 77.31 \\
        \checkmark & ~ & ~ & 59.97 & 82.83 \\ 
        \checkmark & \checkmark & ~ & 61.68 & 84.56  \\ 
        \checkmark & \checkmark & \checkmark & 64.49 & 88.67 \\
    \end{tabular}
    \vspace{-5pt}
    \caption{Ablation Studies on the Mip-NeRF-360 dataset. ``RAD'', ``EL'', and ``FCL'' refer to the proposed Region-Aware Distillation, Entropy Loss, and Feature Consistency Loss, respectively. When the region-aware distillation is not used, we replace it with a standard weighted distillation loss for semantic feature learning.}
    \label{tab:ablation_studies_loss}
    \vspace{-10pt}
\end{table}

\begin{table}[t]
    \centering
    \begin{tabular}{ccc|cc} 
        \multicolumn{3}{c|}{\textbf{Setting}} & \multicolumn{2}{c}{\textbf{Performance}} \\ 
        ID & Method & GaS & \textbf{mIoU(\%)} & \textbf{mAcc(\%)}  \\
        \hline\hline
        a & Langsplat (d=3) & ~ & 54.68 & 73.09 \\
        b & Langsplat (d=3) & \checkmark & 58.27 & 79.59 \\ 
        c & GOI (d=10) & ~ & 58.48 & 68.86 \\ 
        \hline
        d & Ours (d=3) & \checkmark & 60.05 & 85.08 \\
        e & Ours (d=8) & \checkmark & 63.70 & 86.95 \\
        f & Ours (d=16) & \checkmark & 64.49 & 88.67 \\
        g & Ours (d=32) & \checkmark & OOM & OOM \\
    \end{tabular}
    \vspace{-5pt}
    \caption{Ablation studies on feature dimensions and effectiveness of our granularity-aware strategies on the Mip-NeRF-360 dataset.}
    \vspace{-10pt}
    \label{tab:ablation_featdim}
\end{table}


\textbf{Ablation study on losses}.
Additionally, we investigated the impact of our proposed region-aware distillation, entropy loss, and feature consistency loss on performance. \cref{tab:ablation_studies_loss} presents the quantitative evaluation results. We observe that RAD significantly boosts performance, improving mIoU and mAcc by 4.8\% and 5.5\%, respectively. This improvement primarily comes from compensating for the loss weights of small or distant objects, thus enhancing their feature learning. EL reduces feature blurriness by enforcing the feature field to focus on a single scale rather than averaged features. The introduction of FCL further mitigates the impact of outliers, enhancing the stability and consistency of the feature field, resulting in a 2.8\% and 4.1\% improvement in mIoU and mAcc, respectively.

\textbf{Ablation study on feature dimension}.
We also explored the impact of the dimension $d$ of the compressed features stored in Gaussians on the performance of our method. Ablation studies in \cref{tab:ablation_featdim}(d)-(f) show that performance improves as $d$ increases; however, this improvement slows down when $d>8$. At $d=32$, our 24GB VRAM may encounter out-of-memory issues in some complex scenes. Therefore, considering both performance and training cost, we chose $d=16$ as the default setting. 
Comparisons with other approaches shows that our method maintains a performance advantage at the same feature dimension. 
In \cref{tab:ablation_featdim}(b), we further integrated GaS into Langsplat, observing a 3.6\% and 6.5\% improvement in mIoU and mAcc, demonstrating the effectiveness of the proposed GaS module. Additionally, the comparisons in \cref{tab:ablation_featdim}(b) and (d) highlight the advantages of our training-stage granularity selection strategy in GaD over previous inference-stage approaches. 

\section{More Visualization Results}

\begin{figure*}
    \centering
    \vspace{-20pt}
    \includegraphics[width=0.67\textwidth]{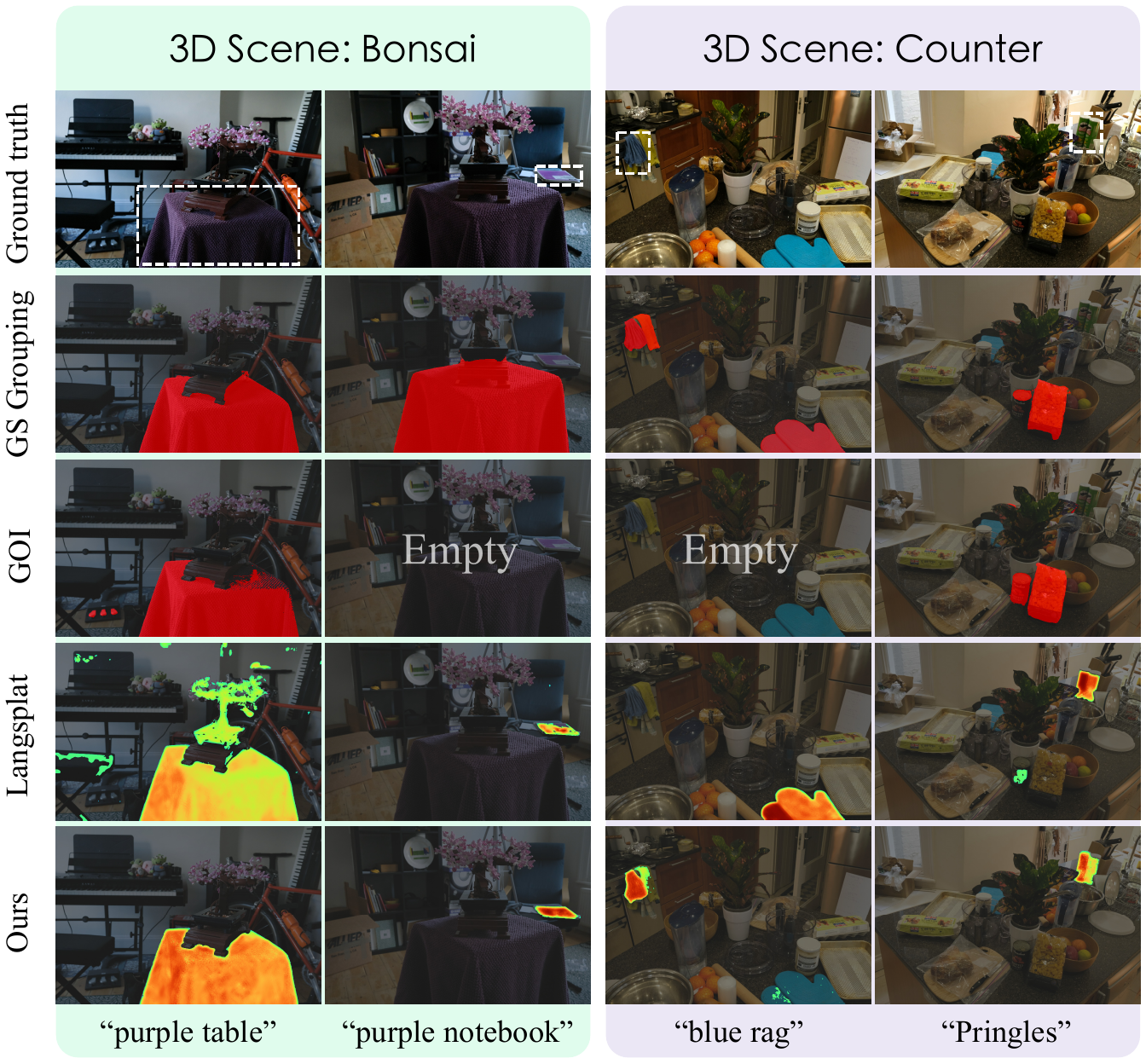}
    \par 
    \vspace{5pt}
    \includegraphics[width=0.67\textwidth]{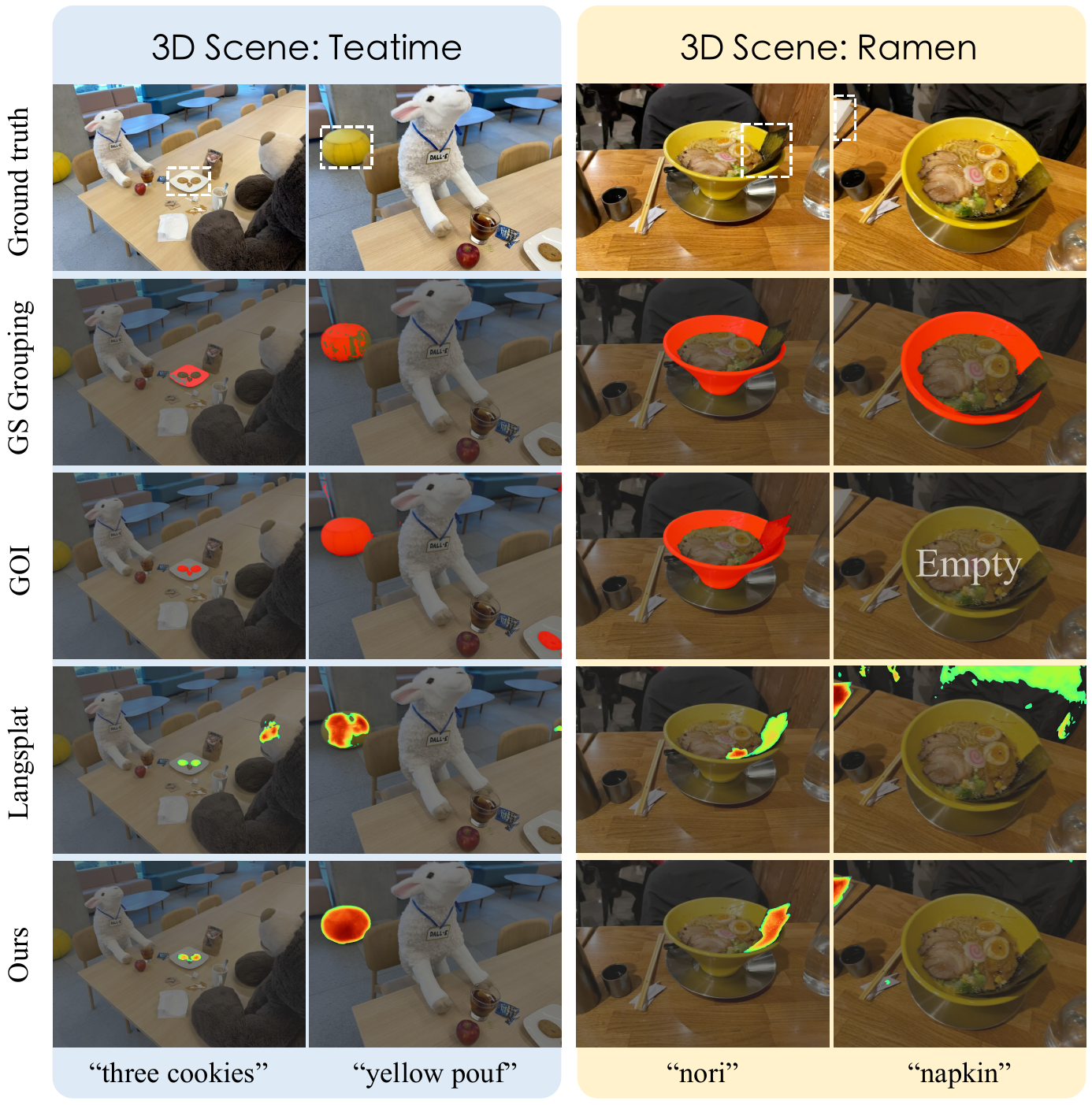}
    \vspace{-5pt}
    \caption{\textit{Additional relevance visualization results of open-vocabulary segmentation}. }
    \label{fig:addtional_comparision}
\end{figure*}

In \cref{fig:addtional_comparision}, we present more open-vocabulary query visualization results on the LERF and Mip-NeRF-360 datasets. Our method demonstrates enhanced stability in complex real-world scenes and superior ability to distinguish semantically similar objects, validating the effectiveness of our proposed granularity-aware learning strategy.

\begin{figure*}
    \centering
    \vspace{-20pt}
    \includegraphics[width=0.77\textwidth]{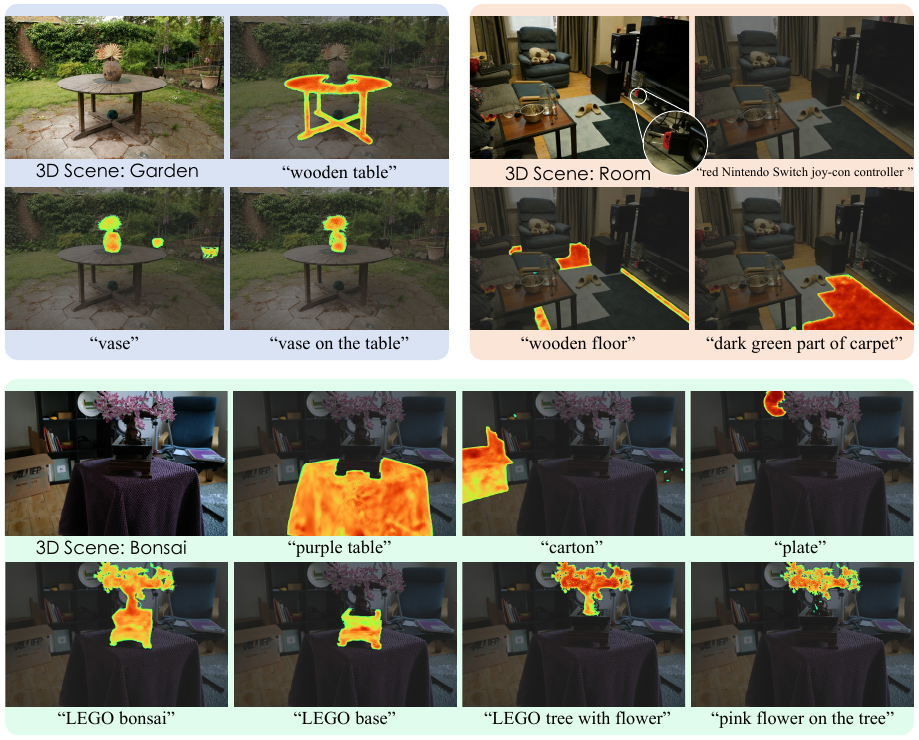}
    \par 
    \vspace{5pt}
    \includegraphics[width=0.77\textwidth]{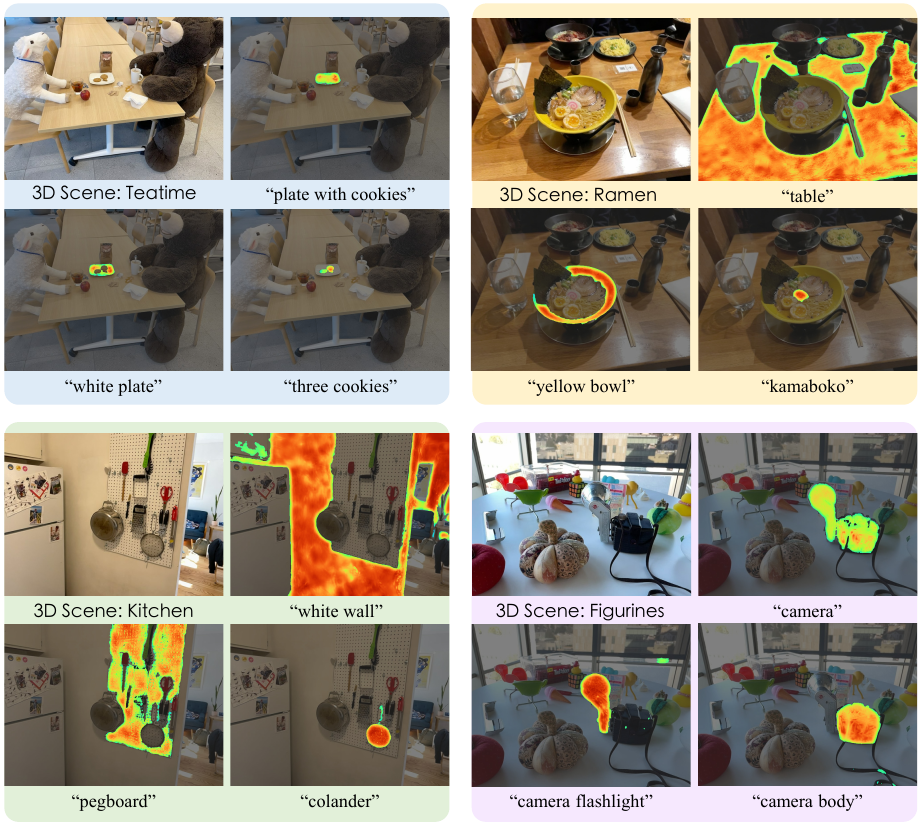}
    \vspace{-5pt}
    \caption{\textit{Additional segmentation results of our GAGS}. Our method is able to produce accurate responses to multi-granularity queries.}
    \label{fig:addtional_example}
\end{figure*}

In \cref{fig:addtional_example}, we visualize additional open-vocabulary semantic segmentation examples of our method on the LERF and Mip-NeRF-360 datasets. Our method is able to accurately responds to queries across various granularities, producing segmented target objects with clear edges.

\section{Limitations}

Although GAGS achieves stable and efficient open-vocabulary scene understanding through its granularity-aware strategy, it still faces several challenges. While the self-supervised granularity optimization during training avoids failure cases in granularity selection caused by outliers during inference, it also leads to some information loss, making it difficult to distinguish part-level objects within complex structures in certain cases. We are actively exploring solutions to this issue to achieve a better balance between performance and efficiency.



\end{document}